\documentclass[10pt,twocolumn,letterpaper]{article}

\usepackage[pagenumbers]{cvpr} 
\usepackage{url}
\usepackage{graphicx}
\usepackage{pgfplots}
\usepackage{pgfplotstable}
\usepackage{algorithm}
\usepackage{algorithmic}
\usepackage{tikz} 
\usepackage{longtable}
\usepackage{xcolor}
\usepackage{colortbl}
\usepackage{booktabs}
\usepackage{multirow}   
\usepackage{array}      
\usepackage{dcolumn}   
\usepackage{fancybox} 
\usepackage[most]{tcolorbox}
\usepackage{siunitx}  
\usepackage{threeparttable} 
\usepackage{adjustbox} 
\usepackage{siunitx}
\usepackage{arydshln}
\usepackage{amssymb}
\pgfplotsset{compat=1.18}

\definecolor{cvprblue}{rgb}{0.21,0.49,0.74}
\usepackage[pagebackref,breaklinks,colorlinks,allcolors=cvprblue]{hyperref}

\def\paperID{15911} 
\def\confName{CVPR}
\def\confYear{2026}

\title{Scaling Spatial Reasoning in MLLMs through Programmatic Data Synthesis}

\begin{document}

\author{
Helu Zhi\textsuperscript{1} \quad Jingjing Huang\textsuperscript{2} \quad Wang Xu\textsuperscript{2\thanks{Corresponding authors}} \quad 
Yangbin Xu\textsuperscript{3} \quad Yibin Huang\textsuperscript{1} \\ 
Wanyue Zhang\textsuperscript{5} \quad Baoyang Jiang\textsuperscript{2} \quad Shirui Deng\textsuperscript{6} \quad Liang Zhu\textsuperscript{6} \\
Fangfang Li\textsuperscript{6} \quad Tiejun Zhao\textsuperscript{1} \quad Yankai Lin\textsuperscript{4*} \quad Yuan Yao\textsuperscript{2*}
\\[0.5em] 
\normalsize \textsuperscript{1} Faculty of Computing, Harbin Institute of Technology \quad \textsuperscript{2} Tsinghua University \\
\normalsize \textsuperscript{3} Institute of Microelectronics of the Chinese Academy of Sciences \quad \textsuperscript{4} Renmin University of China \\
\normalsize \textsuperscript{5}  Institute of Automation, Chinese Academy of Sciences\quad \textsuperscript{6} Central South University \\
\normalsize 
\small zhihelu@126.com, xwjim812@126.com, yankailin@ruc.edu.cn, yaoyuanthu@163.com
}

\maketitle

\begin{figure*}[t!]
    \centering 
    \includegraphics[width=1\linewidth]{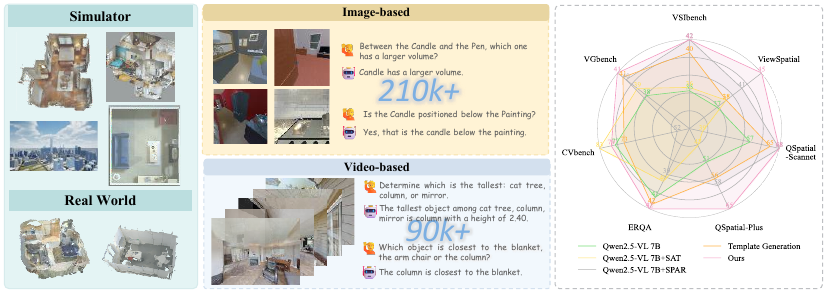} 
    \caption{Overview of SPRITE. Our dataset contains 210k+ image-centered data, 90k+ video-centered data. Models trained with our data have demonstrated improved performance across multiple spatial understanding benchmarks.} 
    \label{fig:overview}
\end{figure*}

\begin{abstract}

Embodied intelligence, a grand challenge in artificial intelligence, is fundamentally constrained by the limited spatial understanding and reasoning capabilities of current models. Prevailing efforts to address this through enhancing Vision-Language Models (VLMs) are trapped in a dilemma: template-based datasets are scalable but structurally rigid, while manual annotation is linguistically diverse but unscalable and, critically, computationally imprecise. We introduce SPRITE, a novel framework that overcomes this dilemma by leveraging simulators and large models to programmatically synthesize scalable, diverse, and high-quality spatial reasoning data. The core innovation of SPRITE is to reframe ground-truth generation as a code-generation task. We utilize LLMs to compile complex spatial questions into executable programs, which are then verified against high-precision scene meta-information extracted from simulators. This ensures our ground truth is both computationally precise and verifiable, while the generative power of LLMs provides vast linguistic diversity. 
Leveraging this pipeline, we have curated a dataset encompassing $3$ simulators, $11k+$ scenes, and $300k+$ image/video instruction-tuning pairs. 
We demonstrate that a VLM trained on our data achieves significant performance gains on multiple spatial benchmarks and outperforms other open-source datasets of equivalent size. Furthermore, a scalability analysis confirms our hypothesis that overcoming the low-diversity nature of traditional template methods is essential for building robust, generalizable spatial intelligence.
We will make the SPRITE framework code and the full $\mathbf{300k+}$ dataset publicly available to facilitate future research in spatial intelligence.\footnote{Code available at \url{https://github.com/AI9Stars/SPRITE}}
\end{abstract}

\section{Introduction}
\label{sec:intro}

Embodied intelligence is a grand challenge in artificial intelligence, aiming to build systems that can interact with and reason about the physical world~\cite{yang2025thinking,cai2025cookbench,wang2023robogen}.
However, a fundamental bottleneck impeding progress is the limited spatial understanding and reasoning capability of current models. 
This capability is paramount, as it transcends simple object recognition~\cite{zhang2025flatland,li2024llava,liu2023visual} and demands a deep, compositional understanding of object relations, 3D poses and scenes. 
Without this robust spatial awareness, models cannot reliably plan, navigate, or manipulate objects in complex, real-world scenarios~\cite{chiang2024mobility,song2024vlm,team2025gemini,zhou2025roborefer}.

To address this deficiency, prevailing research has focused on enhancing Vision-Language Models (VLMs) through specialized datasets~\cite{zhang2025flatland,ray2024sat,chen2024spatialvlm}. 
However, existing data generation paradigms are subject to a trilemma between diversity, scalability, and precision.
Template-based methods~\cite{zhang2025flatland}, while scalable, produce structurally rigid data, severely limiting linguistic variance and failing to capture the combinatorial explosion of complex spatial queries, hus hindering model generalization. Conversely, manual annotation, while capturing linguistic diversity, is not only unscalable but, more critically, computationally imprecise. Human annotators cannot be expected to reliably calculate complex spatiotemporal relationships. 
This reveals a critical gap: the need for a framework that can generate data that is simultaneously scalable, linguistically diverse, and programmatically verifiable in terms of its ground truth.

To overcome this dilemma, we introduce SPRITE, a novel framework that leverages simulators and Large Language Models (LLMs) to programmatically synthesize large-scale, diverse, and high-quality spatial reasoning data. The LLMs are utilized for both diverse question generation and programmatic ground truth acquisition. The core innovation of SPRITE is to reframe the ground-truth generation problem as a code-generation task. Instead of relying on imprecise manual labels, we utilize LLMs to compile complex spatial questions into executable programs. This code is then programmatically executed against high-precision scene meta-information (e.g., object poses, categories, and bounding boxes) extracted from advanced simulators. This approach uniquely ensures that our ground truth is both computationally precise and verifiably correct, while the generative power of LLMs provides vast question diversity. Our pipeline operationalizes this concept: we first collect rich meta-information from $11k+$ scenes, then employ VLLMs to generate a wide array of linguistically diverse spatial questions, and finally, use Code LLMs to generate the executable programs that compute the ground truth.

Leveraging this framework, we have created a comprehensive dataset of $7k+$ videos and $30k+$ images, forming the foundation for our $300k$ instruction-tuning pairs. We demonstrate the efficacy of our approach by training Qwen2.5-VL-7B~\cite{Qwen2.5-VL}, which exhibits significant performance gains on a variety of spatial understanding benchmarks, including VSIbench~\cite{yang2025thinking}, ViewSpatial~\cite{li2025viewspatialbenchevaluatingmultiperspectivespatial}, and SpatialVLM~\cite{chen2024spatialvlm}. Comparative results show that our dataset, SPRITE-300K, outperforms other open-source spatial reasoning datasets of equivalent size. Furthermore, a scalability analysis on scene diversity empirically validates that greater diversity leads to significant gains in model performance. This finding confirms our initial hypothesis: overcoming the rigid, low-diversity nature of traditional template-based methods is essential for building robust, generalizable spatial intelligence.

\section{Related Work}

In recent years, VLMs have achieved remarkable success in cross-modal understanding and reasoning by effectively integrating visual information with natural language. This advancement has led to outstanding performance across applications, including image captioning~\cite{radford2021learning,li2022blip}, visual question answering (VQA)~\cite{antol2015vqa}, and visual navigation~\cite{das2018embodied,wang2023voyager}, enabling models to process complex perceptual information efficiently.

Despite these significant achievements, VLMs still face considerable challenges in spatial understanding. Spatial understanding is critical for models to effectively perceive, comprehend, plan, and act in the real world~\cite{yang2025cambrian,liu2025spatialcot}.
It encompasses understanding their relative spatial relationships and performing complex logical inferences based on these relationships~\cite{yang2025thinking}. Existing VLMs often exhibit limitations when handling tasks requiring precise spatial understanding~\cite{zhang2025mllms}.
This weakness restricts the full potential of VLMs in real-world applications. To address this limitation, current research has pursued several promising directions, primarily focusing on innovations in data augmentation, model architecture, and training methodologies.

\begin{figure*}
    \centering 
    \includegraphics[width=1\linewidth]{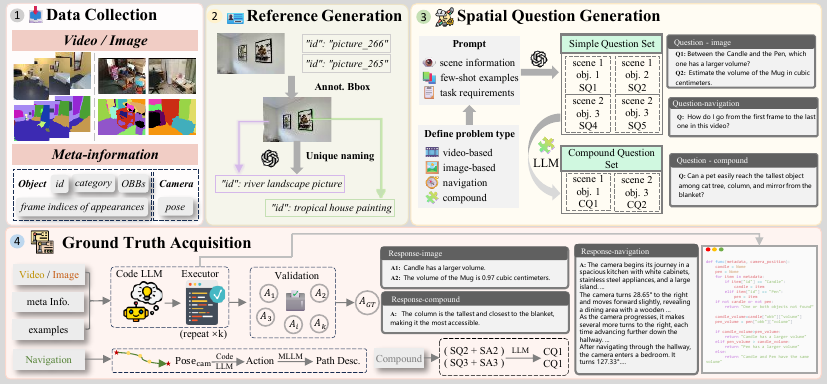} 
    \caption{Data generation approach. Our approach consists of five main parts:(1) Data collection, (2) Reference Generation, (3) Leveraging the collected data to generate diverse questions, (4) Obtaining ground truth from the executable code that processes meta-information. }
    \label{fig:pipeline}
\end{figure*}

From an architectural standpoint, many approaches enhance spatial awareness by explicitly incorporating 3D perception modules into the VLM design. For example, \citet{wu2025spatial} proposes a dual encoder, with a 2D encoder for extracting semantic features and a spatial encoder for extracting 3D result features, which are integrated into a unified visual token through a connector. Similarly,~\citet{zheng2025video} introduces additional modules to convert depth information into 3D coordinates in the global coordinate system to enhance spatial perception. Furthermore, some models integrate point cloud processing capabilities, enabling a more fundamental understanding of 3D geometry and object layouts~\cite{qi2025gpt4scene}. Notwithstanding architectural improvements, the performance of a model is related to abundant data, without which its potential is severely limited.

From a training and algorithmic perspective, reinforcement learning has emerged as a powerful paradigm, as it allows models to learn through interaction and feedback. There are currently many works that enhance the spatial understanding ability of models through reinforcement learning, such as applying the GRPO~\cite{shao2024deepseekmath}. Examples include integrating tool calls for visual drawing to externalize reasoning~\cite{wu2025reinforcing}, or developing a "map imagination" mechanism for navigating unseen environments~\cite{ouyang2025spacer}.
In addition, some works directly apply GRPO to VLM to explore the effects of different training methods~\cite{shen2025vlm,liao2025improved}.
However, the success of these sophisticated training schemes is critically dependent on reliable spatial understanding data. This dependence underscores the fundamental importance of the data generation paradigm itself.

From a data-centric perspective, a primary strategy involves constructing and augmenting datasets specifically designed to foster spatial reasoning~\cite{xu2025multi}. 
Many existing works rely on predefined templates or rules to construct spatial understanding questions and answers. For example, \citet{cheng2024spatialrgpt} utilizes question templates for generating question-answer pairs; \citet{ouyang2025spacer} generates spatial understanding data through rules, particularly suitable for reinforcement learning scenarios. \citet{chen2024spatialvlm} combines human assistance with designed Q\&A templates for data construction . 
\citet{yang2025cambrian}~argues that current Video Multimodal Large Language Models (MLLMs) do not explicitly prioritize space-focused videos during training. To address this, they construct a large-scale, space-focused video dataset based on question templates.

These studies indicate the necessity of large-scale and high-quality data, regardless of whether the focus is on architectural improvements or the design of novel algorithms. However, the template-based method's inherent limitation lies in a lack of diversity~\cite{hudson2019gqa}. The resulting question-answer pairs are often formulaic in phrasing and lacking the richness and generalization ability of natural language. Consequently, we propose \textbf{SPRITE}, a novel framework that resolves the core trade-off in data generation. By reframing ground-truth acquisition as a code-generation task executed against high-precision meta-information, SPRITE is capable of simultaneously generating linguistically diverse questions and computationally precise answers, thereby offering support for research in VLMs.

\section{Methodology}

In this section, we introduce SPRITE, a framework for generating spatial reasoning data via programmatic synthesis. Our pipeline, illustrated in Figure~\ref{fig:pipeline}, consists of four main stages: (1) Data Collection from simulators~(Section~\ref{data_score}), (2) Object Disambiguation via Reference Generation~(Section~\ref{reference_generation}), (3) Diverse Question Generation~(Section~\ref{question_generation}), and (4) Programmatic Ground Truth Acquisition~(Section~\ref{ground_truth}). We conclude by detailing our automated quality control measures~(Section~\ref{quality_check}).

\subsection{Data Collection from Simulators}\label{data_score}

\textbf{Data sourcing and acquisition.}~We leverage simulators including Habitat~\cite{savva2019habitat}, AI2-THOR~\cite{ai2thor}, and AirSim~\cite{airsim2017fsr} to generate synthetic data. These simulators provide object meta-information, including names, oriented bounding boxes (OBBs), and the video frame indices where objects appear.
Within the simulators, we use built-in APIs to acquire information like object names and OBBs. The simulators support multiple camera configurations, allowing us to capture various types of visual data simultaneously. This includes RGB images and semantic segmentation maps. Each pixel in the semantic maps is encoded with a distinct identifier corresponding to its object instance, thereby achieving pixel-accurate instance segmentation. By combining the RGB video stream with the frame-by-frame semantic maps, we can efficiently track each object's appearance and determine its exact frame index.

To enhance the generalizability of our method, we also use the open-source real-world datasets, ScanNet~\cite{dai2017scannet} and ScanNet++~\cite{yeshwanth2023scannet++}. These datasets are meticulously annotated, providing rich object instance IDs and meta-information that can serve as a supplement to our data.

\textbf{Multi-Modal data collection and annotation.}~The proposed framework requires the acquisition of egocentric videos and multi-object scene images from the simulation environment. For video data, we select random start and end points for a camera within the simulator. A path-planning algorithm then generates a movement trajectory. As the agent traverses this trajectory at a predefined constant velocity, we capture RGB and semantic segmentation images at each time step. The captured RGB images are subsequently synthesized to form the final video.

In addition to the video stream itself, we simultaneously collect its meta-information. The meta-information for each video comprises both object information and the camera's pose. Object information consists of an object's ID, category, oriented bounding boxes~(OBBs), and appearance frame indices. The OBBs themselves contain the object's center coordinates, dimensions, and volume. The appearance indices denote the specific frames where an object is visible in the video. Camera pose information includes its 3D position and rotation represented by a quaternion. To reduce computational load, we uniformly sample the video to a fixed length of 32 frames.

The image dataset consists of single-image and multi-image samples. For a single-image sample, we select the top-k frames with the highest object count from the source video. For the multi-image category, a candidate is formed by selecting 2 to 5 images that are arranged chronologically but separated by non-uniform temporal gaps. For each video in our dataset, we extract k candidates following these procedures.
The meta-information for these selected image frames is inherited from the video's meta-information.

\subsection{Reference Generation via VLMs}
\label{reference_generation}
In a given scene, the presence of multiple objects sharing the same name may lead to referential ambiguity when constructing spatial understanding questions. However, directly filtering out these identically named objects would reduce the total number of objects in the scene.

To address object disambiguation, we propose a methodology leveraging reference generation as illustrated in Figure~\ref{fig:pipeline}. For objects of the same category, we select images that collectively cover all instances, providing diverse viewpoints. Based on the semantic segmentation maps, we then annotate each object instance with a distinct color bounding box. These annotated images are then input into GPT-4o~\cite{hurst2024gpt}, which, through its multi-image reasoning, generates a unique name for each. Finally, we map the object aliases back to their respective objects by associating the generated names with the corresponding bounding box colors, and update the scene meta-information with these unique identifiers to ensure consistency in subsequent code generation steps.

\subsection{Diversity Question Generation }\label{question_generation}

\begin{figure}[t!]
    \centering 
    \includegraphics[width=1\linewidth]{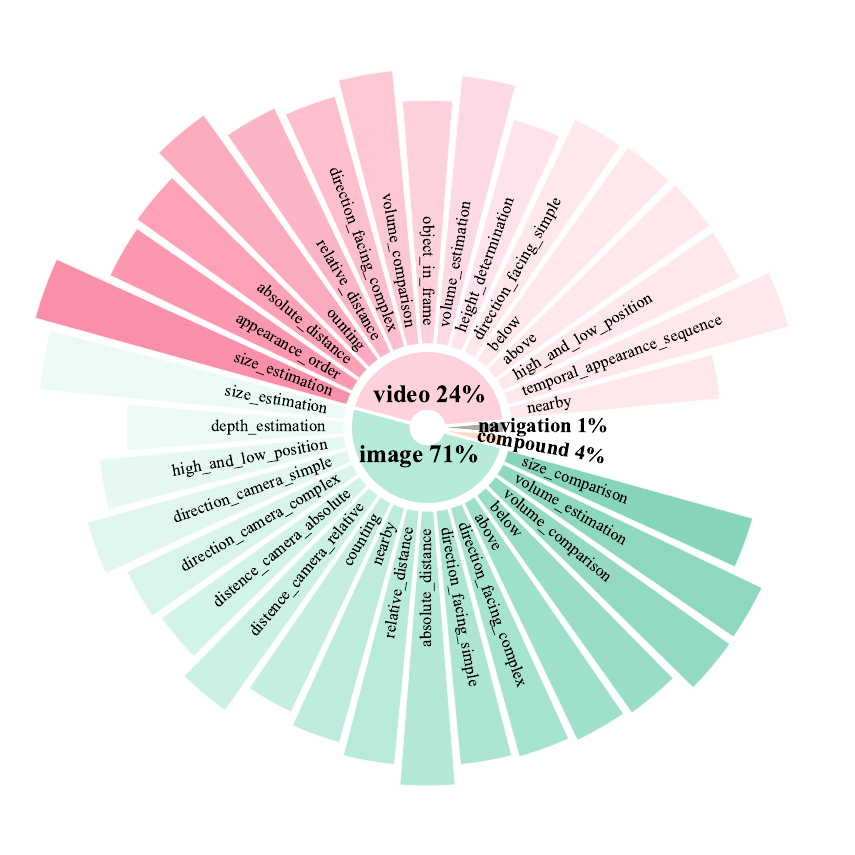} 
    \caption{Spatial task framework. The task framework includes four types of tasks: video-based questions, image-based questions, navigation problems, and compound problems. Among them, the image-based questions contain 18 subclasses, and the video-based questions contain 16 subclasses.}
    \label{fig:task_type}
\end{figure}

\textbf{Spatial task framework.}~To enhance the model's spatial understanding capabilities, we define a set of diverse question types (Figure~\ref{fig:task_type}). For video-based questions, we primarily investigate object relationships, such as size, volume, relative position, and counting, while leveraging the inherent temporal information in videos to define predictive tasks regarding the sequence of object appearances. In image-based questions, the defined problem types encompass not only inter-object relationships (excluding appearance order but otherwise consistent with video questions) but also object-camera relationships, including object depth, distance from the camera, and the prediction of relative object positions from the current viewpoint. This results in over thirty question types covering spatial relationships. Furthermore, we define navigation questions as the description of egocentric camera motion trajectories. To strengthen the VLM's reasoning ability, we construct compound questions formed by combining two questions pertaining to the same object. 

\textbf{Structured question generation.}~We utilize GPT-4o~\cite{hurst2024gpt} to generate video-based questions and image-based questions. Our prompt has three main components: current scene information, few-shot examples, and task requirements. For video-based questions, we provide uniformly sampled frames and all object names and categories. For image-based questions, we provide the corresponding image with object information. Our few-shot examples demonstrate predefined question types, meta-information, and question examples. The task requirements enforce that the generated question must be of a predefined type, involve objects present in the scene, and be answerable using the provided meta-information. The generated output strictly follows the format of the examples, which includes the question, involved objects' names and categories, and question type. This structured output (Figure~\ref{fig:pipeline}, Question Generation section) is essential for automatically parsing the questions and obtaining the ground truth.

\textbf{Compound question synthesis.}~We define compound questions as the combination of two questions related to the same object. Specifically, we hypothesize that a compound question can be decomposed into two questions.
As illustrated in Figure~\ref{fig:compound}, we leverage the reasoning capabilities of LLMs by providing a prompt that includes two questions containing the same object and an example. The model then synthesizes a new compound question based on these inputs.

\subsection{Ground Truth Acquisition}\label{ground_truth}

\textbf{Ground truth from executable code.}~For video-based questions and image-based questions, we leverage LLMs' reasoning capabilities to generate executable code for obtaining ground truth from the provided meta-information. Specifically, we employ a Code LLM(e.g., Qwen3-32B~\cite{qwen3}) to generate code that can process the meta-information. The prompt consists of question information, few-shot examples, and task requirements. The question information includes not only the question itself but also details about the objects and their categories involved. Crucially, to prevent the LLM from hallucinating or hard-coding answers, the provided few-shot examples encourage the generation of generic algorithms that query the meta-information variable programmatically, rather than relying on direct value injection. The few-shot examples contain sample code and sample meta-information. The task requirements are to generate a piece of code that can process the meta-information, with the output of the code being a string. To improve the code's execution accuracy, we also require that the objects mentioned in the code must be the same as the objects involved in the question. The generated code encapsulates the logic to derive the answer, and by executing it, we automatically obtain the ground truth for each question (Figure~\ref{fig:pipeline}, Ground Truth Acquisition section).

\textbf{Navigation ground truth generation.}~Instead of generating low-level control actions, we focus on the model's ability to comprehend and describe egocentric motion. The inherent complexity of autonomous navigation problems necessitates precise and dynamic state representation. First, camera pose information (position and rotation angles) is computationally processed to generate a structured data format. Subsequently, we leverage the advanced capabilities of the GPT-4o~\cite{hurst2024gpt} to convert this structured description into a more intuitive and fluent natural language expression.

\begin{figure}[t!]
\begin{center}
\begin{tcolorbox}[colback=gray!5!white,colframe=gray!75!black,title=Compound Question]
\textbf{Question A:}
Determine which is the tallest: the cat tree, column, or mirror.

\textbf{Answer A:}
The tallest object among the cat tree, column, and mirror is the column, with a height of 2.4 meters.

\vspace{1em} 

\textbf{Question B:}
Which object is closest to the blanket, the armchair or the column?

\textbf{Answer B:}
The column is closest to the blanket.

\vspace{1em}

\textbf{Compound Question:}
Can a pet easily reach the tallest object among the cat tree, column, and mirror from the blanket?

\textbf{Compound Answer:}
The column is the tallest and also the closest to the blanket, making it the most accessible.

\end{tcolorbox}
\end{center}
\caption {\label{fig:compound}Compound question example.}
\end{figure}

\textbf{Compound question ground truth synthesis.} 
As described in Section 3.3, each compound question is synthesized from two constituent sub-questions. To generate its ground truth, we first acquire the answers to both sub-questions by employing a code-generating LLM (code-llm) to generate and execute code. Subsequently, we provide these two resulting question-answer pairs as context to an LLM, leveraging its reasoning capabilities to synthesize the final, composite answer for the compound question. As illustrated in Figure~\ref{fig:compound}, this process enables the model to infer the correct answer by reasoning upon the two pre-existing pairs. The logical validity of this synthesis is subsequently rigorously verified in our quality control stage.

\subsection{Automated Quality Control}
\label{quality_check}
To ensure data quality, we implement rigorous verification procedures. Initially, we systematically filtered the problem set by applying a series of predefined exclusion criteria. For instance, we excluded problem instances where the objects involved were absent from the meta-information list, or where code execution resulted in an error or the inability to locate the specified objects. Moreover, we adopt a voting-based validation approach: for each question, we instruct code LLMs (Qwen3-32B) to generate $k=3$ distinct programs for the same question by using diverse prompts. For the execution results of each candidate code, we use LLMs to determine whether the results are consistent. Inconsistent answers lead to the question being filtered out. 

For compound questions, we employ LLMs to assess whether the final compound answer is a logically consistent inference based on the programmatically verified answers of its constituent sub-questions. Instances where the final answer is deemed implausible or logically inconsistent with these premises are filtered out, thereby enhancing the reliability of the dataset.

\begin{table*}[t!]

   \centering

   \small 
   \caption{Comparison of different visual-spatial language models on various benchmarks.}  \label{tab:main_results}
   \setlength{\tabcolsep}{4pt}

   \begin{tabular}{lcccccccc} 
    \toprule
     Model & VSIbench & ViewSpatial & QSpatial-Scannet & QSpatial-Plus & ERQA & CVbench & VGbench & overall \\
    \midrule
    
    \rowcolor{lightgray} 
    \multicolumn{9}{l}{\textit{Close-source Models}} \\
    GPT-4o & 34.00* & \textbf{38.32*} & 69.41 & 43.56 & \textbf{47.00*} & 64.33 & 40.10 & 48.30\\
    Claude-3.7-Sonnet & \textbf{47.00}* &  34.16 & \textbf{72.35} & \textbf{74.26} & 37.50* & 63.56 & 43.60 & 53.24 \\
    Gemini-2.0 Flash & {45.50} & 36.75* & 55.45 & 71.76 & 46.30* & \textbf{80.53} & \textbf{46.90} & \textbf{54.74} \\
    
    \cmidrule(lr){1-9} 
    \rowcolor{lightgray} \multicolumn{9}{l}{\textit{Open-source Models Trained on Open Spatial Understanding Datasets}} \\ 
    InternVL3-8B & 43.19 & 41.90 & 58.24 & 47.52& \textbf{40.75} & 81.22 & 34.97 & 49.68\\
    \quad+\textbf{SPRITE}~(ours) & \textbf{46.31} & \textbf{44.00} & \textbf{61.76} & \textbf{55.45} & 38.00 & \textbf{82.34} & \textbf{36.17} & \textbf{52.00} \\ 
    \cmidrule(lr){1-9} 
    Qwen2.5-VL 7B & 35.30 & 38.44 & 57.06 & 51.49 & 41.25 & 74.55 & 37.90 & 48.00 \\
    \quad+SAT~\cite{ray2024sat} & 35.74 & 39.95 & 38.82 & 46.53 & 39.50 & 78.51 & 38.98 & 45.43 \\
    \quad+SPAR~\cite{zhang2025flatland} & \textbf{42.07} & 43.79 & 65.29 & 58.42 & 38.75 & 77.12 & 40.30 & 52.25 \\ 
    
    \quad+\textbf{SPRITE}~(ours) & 42.02 & \textbf{44.61} & \textbf{68.24} & \textbf{65.35} & \textbf{42.50} & \textbf{78.58} & \textbf{41.32}  & \textbf{54.66} \\
 
    \bottomrule
    \end{tabular}
   
\end{table*}

\section{Experiments}

\subsection{Experiment Setups}
\label{sec4}

\textbf{Dataset.}~We have constructed and collected $7k+$ videos and $30k+$ images from both a simulator and open-source datasets. Using our proposed approach, we built a spatial understanding dataset containing $300k$ examples. 
Our dataset covers various aspects of spatial reasoning, including $90k$ video spatial reasoning data and $210k$ image spatial reasoning data. The $90k$ video spatial reasoning data, for instance, is further categorized into $75k$ examples based on the task framework, $12k$ compound reasoning questions, and $3k$ instances of navigation data.

\textbf{Model Training.}~We use Qwen2.5-VL-7B~\cite{Qwen2.5-VL} as our baseline model. 
The model was fine-tuned for one epoch using LoRA~\cite{hu2022lora} with a rank of 256 and a learning rate of $5 \times 10^{-6}$. 
For video input, a uniform frame sampling strategy was employed, fixing the number of sampled frames at 32. The specifications of the training parameters are provided in Appendix~\ref{hyperparameter}.

\textbf{Evaluation Benchmarks.}~To assess the efficacy of our constructed dataset, we conducted evaluations on multiple spatial understanding benchmarks, including VSIbench~\cite{yang2025thinking}, ViewSpatial~\cite{li2025viewspatialbenchevaluatingmultiperspectivespatial}, QSpatialBench-ScanNet, QSpatialBench-Plus~\cite{liao2024reasoningpathsreferenceobjects}, ERQA~\cite{team2025gemini}, CVbench~\cite{tong2024cambrian1}, VGbench~\cite{wu2025spatialscore}. 
To ensure a fair comparison that reflects the scope of our dataset, we made two adjustments to the evaluation protocols.
We excluded object-grounding questions from CVbench, as our dataset does not target the grounding task.
Moreover, we omitted the `object view orientation' subtask of ViewSpatial because the object-orientation metadata contains substantial errors.
Detailed descriptions of the benchmarks are provided in Appendix~\ref{benchmark}.

\textbf{Compared Methods.}
To validate the effectiveness of our proposed approach, we conducted a comparative analysis. We benchmarked our model against baseline models that utilize two alternative spatial data construction methods: SAT~\cite{ray2024sat} and SPAR~\cite{zhang2025flatland}. To ensure a fair comparison, all models were trained under identical conditions, including a data volume of $300k$ samples and the same hyperparameter settings. SPAR generates single-view and multi-view data based on a set of predefined task templates, using data that is annotated with 3D grounding information. SAT generates static and dynamic data interactively within a simulator, also following defined task templates. Furthermore, to situate our model's performance within the broader landscape of state-of-the-art (SOTA) systems, we also evaluated it against several leading proprietary models, namely GPT-4o~\cite{hurst2024gpt}, Claude-3.7-Sonnet~\cite{anthropic2024claude37sonnet}, and Gemini-2.0 Flash~\cite{team2025gemini}, on the same set of benchmarks. What's more, to assess the generalizability and transferability of our dataset, we extended our training and validation experiments to InternVL3-8B~\cite{zhu2025internvl3}.

\begin{figure*}[t!] 
    \centering
    
    \begin{subfigure}[t]{0.32\textwidth} 
        \centering
        \begin{tikzpicture}[scale=0.7, transform shape]
        \begin{axis}[
            xlabel={Scene scale~($k$)},
            ylabel={Score~(\%)},
            xmin=0.05, xmax=10,
            xmode=log,
            ymin=40, ymax=56,
            xtick={0.063,0.125, 0.25, 0.5, 1, 2, 4,8},
            xticklabels={1/16,1/8,1/4,1/2,1,2,4,8},
            grid=major,
            major grid style={dotted, gray!50},
            legend style={
                at={(0.5,-0.2)}, 
                anchor=north,
                legend columns=2, 
                draw=none,
                fill=none,
                font=\small
            },
            label style={font=\small},
            tick label style={font=\small},
            title style={font=\normalsize}
        ]
        
        \addplot[
            red!80!black, 
            mark=*,
            mark size = 2pt,
            mark options={fill=red!80!black, draw=red!80!black},
            only marks
        ] coordinates {
            (0.063,45.07)
            (0.125, 44.72)
            (0.25, 48.10)
            (0.5, 51.76)
            (1, 52.74)
            (2, 52.16)
            (4, 52.36)
            (8, 53.36)
        };
        \addlegendentry{SPRITE}

        \addplot[
            blue!80!black,
            mark=triangle*,
            mark size = 2pt,
            mark options={fill= blue!80!black, draw= blue!80!black},
            only marks
        ] coordinates {
            (0.063,41.53)
            (0.125, 42.92)
            (0.25, 46.71)
            (0.5, 47.9)
            (1, 49.10)
            (2, 50.74)
            (4, 51.98)
            (8,51.58)
        };
        \addlegendentry{Template}

        \addplot[
            domain=0.063:8, 
            samples=100,
            red!80!black,
            line width=2pt,
            dashed
        ] {55.0799 -3.5829 * (x^(-0.4083))}; 

        \addplot[
            domain=0.063:8, 
            samples=100,
             blue!80!black,
            line width=2pt,
            dashed
        ] {55.0379 -5.6853 *(x^(-0.3238))}; 
        
        \end{axis}
        \end{tikzpicture}
        \caption{Scaling laws for the number of scenes.}
        \label{fig:scaling_scene} 
    \end{subfigure}%
    \hfill%
    \begin{subfigure}[t]{0.32\textwidth} 
        \centering
        \begin{tikzpicture}[scale=0.7, transform shape]
        \begin{axis}[
            xlabel={Number of training samples~($w$)},
            ylabel={Score~(\%)},
            xmin=2, xmax=31,
            ymin=48, ymax=56,
            xtick={0,3,6,9,12,15,18,21,24,27,30},
            xticklabels={0,3,6,9,12,15,18,21,24,27,30},
            grid=major,
            major grid style={dotted, gray!50}, 
            legend style={
                at={(0.5,-0.20)}, 
                anchor=north,
                legend columns=2,
                draw=none,
                fill=none,
                font=\small
            },
            label style={font=\small},
            tick label style={font=\small},
            title style={font=\normalsize}
        ]
        \addplot[
            red!80!black, 
            mark=*, 
            mark size = 2pt,
            only marks
        ] coordinates {
            (3, 49.173)
            (6, 52.124)
            (9, 53.435)
            (12, 52.889)
            (15, 53.11)
            (18, 53.839)
            (21, 54.489)
            (24, 55.01)
            (27, 54.21)
            (30, 54.567)
        };
        \addlegendentry{SPRITE}

        \addplot[
            blue!80!black, 
            mark=triangle*, 
            mark size = 2pt,
            mark options={fill=blue!80!black, draw=blue!80!black},
            only marks
        ] coordinates {
            (3, 49.38)
            (6, 50.35)
            (9, 52.74)
            (12, 52.805)
            (15, 52.467)
            (18, 52.125)
            (21, 50.752)
            (24, 51.014)
            (27, 50.738)
            (30, 50.611)
        }; \addlegendentry{Template}

        \addplot[
            domain=3:30, 
            samples=100,
            red!80!black,
            line width=2pt,
            dashed
        ] {59.2847-15.3055  * (x^(-0.3607))};

        \addplot[
            blue!80!black,
            dashed,
            domain=3:30,
            samples=100,
            line width=2pt,
            no marks,
        ] {0.00136642*x^3-0.07991518*x^2+1.33528497*x + 45.74636667};

        \end{axis}
        \end{tikzpicture}
        \caption{Comparison of our method and the template.}

        \label{fig:scaling_count}
    \end{subfigure}%
    \hfill%
    \begin{subfigure}[t]{0.32\textwidth} 
        \centering
        \begin{tikzpicture}[scale=0.7, transform shape]
        \begin{axis}[
            xlabel={Number of training samples~($w$)},
            ylabel={Score~(\%)},
            xmin=2, xmax=31,
            ymin=48, ymax=56,
            xtick={0,3,6,9,12,15,18,21,24,27,30},
            xticklabels={0,3,6,9,12,15,18,21,24,27,30},
            grid=major,
            major grid style={dotted, gray!50}, 
            legend style={
                at={(0.5,-0.2)}, 
                anchor=north,
                legend columns=3, 
                draw=none,
                fill=none,
                font=\small
            },
            label style={font=\small},
            tick label style={font=\small},
            title style={font=\normalsize}
        ]

        \addplot[
            red!80!black, 
            mark=*, 
            mark size = 2pt,
            only marks
        ] coordinates {
            (3, 49.173)
            (6, 52.124)
            (9, 53.435)
            (12, 52.889)
            (15, 53.11)
            (18, 53.839)
            (21, 54.489)
            (24, 55.01)
            (27, 54.21)
            (30, 54.567)
        };\addlegendentry{Hybrid}

       \addplot[
            green!60!black, 
            mark=square*,
            dash pattern=on 5pt off 3pt on 1pt off 3pt, 
            mark size = 2pt,
            only marks
        ] coordinates {
            (3, 48.93)
            (6, 51.05)
            (9, 53.07)
        };\addlegendentry{Video}

        \addplot[
            blue!80!black, 
            mark=triangle*, 
            mark size = 2pt,
            only marks
        ] coordinates {
            (3, 49.60)
            (6, 51.51)
            (9, 52.28)
            (12, 52.91)
            (15, 52.38)
            (18, 52.12)
            (21, 53.51)
        };\addlegendentry{Image}
        \addplot[
            domain=3:21, 
            samples=100,
            blue!80!black, 
            line width=2pt,
            dashed
        ] {53.5100  -11.4827 * (x^((-0.9834))}; 

        \addplot[
            domain=3:30, 
            samples=100,
            red!80!black,
            line width=2pt,
            dashed
        ] {59.2847-15.3055  * (x^(-0.3607))};

        \end{axis}
        \end{tikzpicture}
        \caption{The effectiveness of image and video data.}
        \label{fig:scaling_vi}
    \end{subfigure}%
    \caption{Comprehensive analysis of data scaling and model performance. (a) Scaling laws with respect to the scene scale ($k$). (b) Performance comparison between our method and the template across varying numbers of training samples ($w$). (c) Ablation study on different data modalities (image, video, and hybrid) for the number of training samples ($w$).}
\end{figure*}
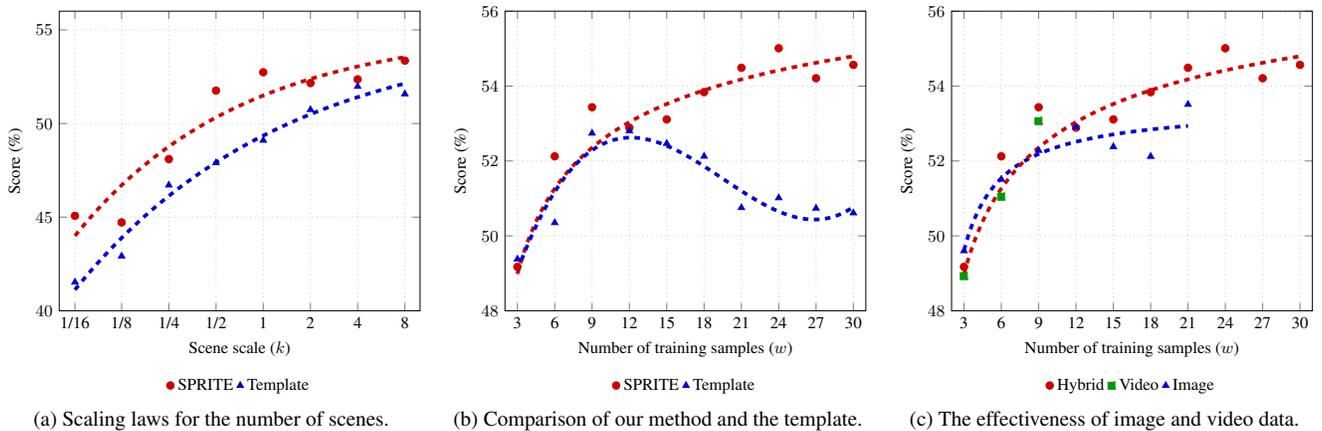

\subsection{Main Results}

The main results, presented in Table~\ref{tab:main_results}, underscore the effectiveness of our proposed dataset. Training the Qwen2.5-VL-7B model on our dataset yields substantial performance gains across all seven evaluated spatial understanding benchmarks. Notably, the improvements are particularly pronounced on VSIbench, QSpatialBench-ScanNet, and QSpatialBench-Plus, with performance increases exceeding 5\%. Furthermore, the model achieves gains of over 3\% on benchmarks such as ERQA and View-Spatial. These results provide compelling evidence that our dataset and construction methodology effectively enhance the model's spatial reasoning capabilities.

When compared with proprietary closed-source models, the Qwen2.5-VL-7B trained on our dataset demonstrates highly competitive performance. Specifically, its scores on VSIbench, QSpatial-Scannet, and QSpatial-Plus approach Claude-3.7-Sonnet. On CVBench, our model's performance is comparable to that of Gemini-2.0 Flash, highlighting the validity of our data.

In a direct comparison with models trained on existing public datasets, our approach shows a distinct advantage. The model trained on our dataset achieved a superior overall score, outperforming models trained on SPAR (52.25) and SAT (45.43). This outcome indicates that our proposed data construction method is more effective than those used for SAT and SPAR.

To further validate the generalizability of our method, we extended our experiments to a distinct model architecture, InternVL3-8B. Following training on our dataset, the model's overall performance increased from its baseline of 49.68 to 52.00. This improvement confirms that our proposed dataset is not model-specific and can serve as a valuable resource for enhancing spatial reasoning in a variety of vision-language models.


\begin{table*}[t!]
   \caption{Comparison of code and natural language~(NL).}
   \label{tab:code_direct}
   \centering
   \small 
   \setlength{\tabcolsep}{4pt} 
  
   \begin{tabular}{lcccccccc} 
    \toprule
     Model & VSIbench & ViewSpatial & QSpatial-Scannet & QSpatial-Plus & ERQA & CVbench & VGbench & overall \\
    \midrule
    Qwen2.5-VL 7B & 35.30 & 38.44 & 57.06 & 51.49 & 41.25 & 74.55 & 37.90 & 48.00\\
    \quad+NL  & 38.77 & \textbf{39.59} & \textbf{58.24} & 51.49  & 41.75  & \textbf{76.29}  & 39.43  & 49.36  \\  
    \quad+SPRITE~(ours) & \textbf{38.91} & 39.44 & \textbf{58.24} & \textbf{56.44}  & \textbf{42.50}  & \textbf{76.29}  & \textbf{39.65} & \textbf{50.21}  \\    
    \bottomrule
  \end{tabular}

\end{table*}

\subsection{Analysis Experiment}
In this section, we systematically designed and conducted a series of experiments, leveraging Qwen2.5-VL-7B as the base model, to investigate the impact of scene diversity, template-based data construction methods, training data type, code generation method, and model training strategies on performance.
The training hyper-parameter is consistent with the main experiments as described in Section~\ref{sec4}.

\textbf{Scalability of the number of scenes.}
To verify the data diversity from the perspective of the scene, we conducted a scalability experiment based on the number of scenes. We logarithmically increase the number of scenes up to $8k$. To further analyze the differences between our method and the template method, we conducted experiments on the template data with the same settings. For the template method, based on the task framework defined by the section~\ref{question_generation}, we designed question templates and manually wrote code for each question template to obtain the ground truth.

The experimental results are shown in Figure~\ref{fig:scaling_scene}. The results demonstrate a clear positive correlation between the number of scenes and the overall score. Specifically, the performance exhibits a rapid improvement with an increase in the number of scenes at low numbers. As the number of scenes continually increases, the rate of improvement significantly decreases, following the principle of diminishing returns. This finding suggests that data diversity is a critical factor for performance improvement.

Furthermore, both our method and the baseline exhibit a similar scaling trajectory, confirming that increasing scene variety is a fundamentally effective strategy regardless of the underlying data generation technique. 

More importantly, our method maintains a consistent performance margin over the baseline across the entire spectrum of scene counts. This consistent superiority demonstrates the enhanced quality and effectiveness inherent to our data generation strategy, validating its ability to produce more valuable training data.

\textbf{Comparison of our method and the template.}
We investigated the scalability of our method by comparing it with the template-based approach. 
We fixed the number of unique scenes at $6k$ and scaled the data volume incrementally to a maximum of $300k$.

As shown in Table~\ref{fig:scaling_count}, the two methods exhibit different scalability profiles. The performance of the template-based method is highly sensitive to data volume, showing a degradation as the dataset size grows beyond $120k$. This suggests a critical limitation: as more data is generated from a fixed set of templates, the diminishing diversity and potential decline in quality severely hamper performance. The performance degradation observed in the template-based method beyond 120k samples is likely due to overfitting to the rigid syntactic patterns inherent in templates. This excessive exposure to repetitive structures harms the model's ability to generalize to the diverse linguistic formulations found in the evaluation benchmarks.

In stark contrast, our method demonstrates remarkable robustness, maintaining a high and stable level of performance across the entire data range. This result underscores the effectiveness of our proposed data generation strategy, which ensures data quality and diversity, thereby achieving excellent scalability where template-based methods falter.

\textbf{The effectiveness of image and video.}
To investigate the distinct contributions of image and video modalities, we designed a comprehensive data scaling study. We constructed three distinct data pools: image-only, video-only, and a hybrid combination. From these pools, we created training sets of increasing sizes, starting from $30k$ up to $300k$ samples in increments of $30k$, to analyze performance trends and scaling behaviors. 

Our experiments demonstrate that hybrid training on both images and videos yields superior performance compared to training on either modality alone, as shown in Figure~\ref{fig:scaling_vi}. This indicates that the hybrid data strategy effectively leverages the complementary nature of image and video information. 

When training with video data alone, the model can perform and grow steadily. Therefore, the video data has played a positive role. Training on images alone achieves strong results, but its performance plateaus and is eventually surpassed by the hybrid approach as data volume increases, underscoring the limited generalization capability of models trained exclusively on images.

\textbf{Validation of code-based ground truth generation.} 

We construct our ground truth dataset by programmatically generating solutions using a code-based reasoning model. To evaluate the superiority of our programmatic data synthesis, we created a test set of $20k$ samples from $11k$ scenes and compared our method against a baseline: direct natural language (NL) inference performed by the same model Qwen3-32B~\cite{qwen3}. The NL baseline was prompted with the question and full textualized meta-information to infer the answer directly.

As shown in Table $\ref{tab:code_direct}$. The code-based approach demonstrates a discernible advantage over direct NL inference, achieving higher scores on benchmarks like VSIbench, ERQA, and VGbench. This superiority substantiates the efficacy of the code generation paradigm in providing a more accurate and verifiable ground Truth for complex spatial reasoning problems. 

While the comparable overall performance suggests that NL direct inference retains reasonable accuracy for simpler queries, its reliance on longer context chains introduces increased time and computational overhead, limiting its ability to scale to the complexity. Thus, this affirms the code-based method as a more scalable and robust solution for high-quality dataset construction.

\begin{table}[t!]
\centering
\small 
\caption{The effect of different training paradigms.}
\label{tab:sft_gppo}
\setlength{\tabcolsep}{4pt} 
\begin{tabular}{l c c c c}
\toprule
Method & NUM & VSIbench & CVbench & VGbench \\
\midrule
Qwen2.5VL-7B & {-} & 35.30 & 74.55  & 37.90 \\
\quad+SFT &6.4k & 36.18 & 75.93  & 38.00 \\
\quad+GRPO & 6.4k & \textbf{39.47} & \textbf{79.26}  & \textbf{39.02} \\
\bottomrule
\end{tabular}

\end{table}

\textbf{The effect of different training paradigms.}
To investigate the characteristics of our constructed dataset under different training paradigms, we conducted a comparative study utilizing both Supervised Fine-Tuning (SFT) and a representative Reinforcement Learning (RL) algorithm, GRPO. We sampled $6.4k$ data points and converted them into a multiple-choice question format by using an LLM to generate plausible distractors. This conversion is critical for RL, as it facilitates the generation of clear, verifiable reward signals based on the correctness of the model's responses.

As presented in Table \ref{tab:sft_gppo}, with a dataset size of $6.4k$, the model performance achieved through GRPO surpasses that of SFT. This outcome suggests that GRPO is capable of optimizing the model more effectively by employing efficient reward signaling mechanisms to leverage sparse data. Consequently, GRPO achieves marked performance gains even when the available data is limited. Meanwhile, this proves that the data constructed by our method can meet the requirements of different training methods.

\section{Conclusion}

In this paper, we tackled the critical deficiency of current VLMs in spatial understanding. 
We argue that existing datasets, which rely on templates or manual annotation, suffer from fundamental limitations in diversity, scalability, and cost. 
To overcome this, we introduce SPRITE, a novel framework that leverages simulators and LLMs to automatically construct a high-quality, large-scale spatial reasoning dataset. 
Our core contribution is that we use code LLMs to produce executable code for diversity questions, which is then used to derive ground-truth answers via precise spatial computation. 
This approach eliminates reliance on rigid templates and ensures high accuracy. 
Using this method, we generated a comprehensive dataset with over $300k$ instruction-following pairs.
Experiments validate the efficacy of our method. Models trained on our data demonstrate significant performance improvements across multiple benchmarks, consistently outperforming those trained on template-based datasets. These results confirm that our proposed framework is an effective solution to enhance the spatial capabilities of VLMs.
{
    \small
    \bibliographystyle{ieeenat_fullname}
    \bibliography{main}
}


\clearpage
\appendix

\newpage
\section{Appendix}
\subsection{Simulator}\label{simulator_info}
Our research primarily uses simulators such as Habitat-Sim, AI2-Thor, and AirSim to collect data. The following is a brief introduction to these simulators.

\textbf{Habitat-Sim.}~Habitat-Sim is a high-performance 3D simulator that supports physics engines. It is designed for 3D spatial scans of both indoor and outdoor environments and includes built-in support for datasets like HM3D and MP3D. The simulator can be configured with various sensors, including RGB-D cameras and semantic maps. Additionally, it features collision detection for robots as they move, making it a popular choice for embodied AI experiments such as navigation and instruction following.

\textbf{AI2-Thor.}~AI2-THOR is a nearly photorealistic and interactive framework for embodied AI agents. In addition, ProcTHOR~\cite{procthor} can be run under the AI2-THOR framework, which includes 11k scenes. Meanwhile, it uses procedural generation to sample massively diverse, realistic, interactive, customizable, and performant 3D environments.

\textbf{AirSim.}~AirSim is an open-source simulator for cars and drones built on Unreal Engine. It supports multiple outdoor environments, such as the Embodied City, and provides APIs that allow you to control a drone's movement as if you were playing a game. The simulator also supports RGB-D sensors.

\subsection{Spatial Task Framework}\label{task_define}
There are image-related question types 
\begin{itemize}
    \item \textbf{object\_size\_estimation.}~Given an object, identify the size of the object and output the specific value.
    \item \textbf{object\_size\_comparison.}~Given multiple objects, compare their sizes.
    \item \textbf{object\_volume\_estimation.}~Identify the volume of the object and output the corresponding value.
    \item \textbf{object\_volume\_comparison.}~Compare the volume sizes of multiple objects in the video to find the largest or smallest one. 
    \item  \textbf{object\_below.}~Determine whether one object is positioned at a lower elevation relative to the other.
    \item \textbf{object\_above.}~Assess the relative vertical positions of two objects to establish whether one is positioned above the other.
    \item \textbf{object\_direction\_facing\_complex.}~Given three objects, after determining the perspective through two of them, determine the relative position of the other object in the video. Relative position includes front-left, front-right, back-left, or back-right.
    \item \textbf{object\_direction\_facing\_simple.}~Given three objects, after determining the perspective through two of them, determine the relative position of the other object in the video. Relative position includes left, right. 
    \item \textbf{object\_absolute\_distance.}~Calculate the absolute distance between two objects in the image.
    \item \textbf{object\_relative\_distance.}~Determine which of the candidate objects is the closest to the target object.
    \item \textbf{object\_nearby.}~Determine if there are objects within a certain range near a specified object.
    \item \textbf{object\_counting.}~Count the total number of times a certain type of object appears in the image.
    \item \textbf{object\_distance\_camera\_relative}~From the perspective of the current picture, identify which of the two objects is farther or closer to the camera
    \item \textbf{object\_distance\_camera\_absolute.}~From the perspective of the current picture, identify the distance from an object to the camera's perspective.
    \item \textbf{object\_direction\_camera\_complex.}~From the current perspective of the picture, determine the positional relationship between two objects. Relative position includes front-left, front-right, back-left, or back-right. 
    \item \textbf{object\_direction\_camera\_simple.}~From the current perspective of the picture, determine the positional relationship between two objects, whether one object is to the left or right of the other.
    \item \textbf{high\_and\_low\_position.}~Given two objects, assess their relative vertical positions to identify which one is at a higher elevation.
\end{itemize}

There are video-related question types:

\begin{itemize}
    \item \textbf{object\_appearance\_order.}~Identify the order in which objects appear in the video and list the object names in chronological order.
    \item \textbf{object\_absolute\_distance.}~Calculate the absolute distance between two objects in the video. 
    \item \textbf{object\_counting.}~Count the total number of times a certain type of object appears in the video and output this value. 
    \item \textbf{object\_relative\_distance.}~Determine which of the candidate objects is the closest to the target object.
    \item \textbf{object\_size\_estimation.}~Determine the length, width, and height of the objects in the video and output the corresponding dimensions.
    \item \textbf{object\_direction\_facing\_complex.}~Given three objects, after determining the perspective through two of them, determine the relative position of the other object in the video. Relative position includes front-left, front-right, back-left, or back-right. 
    \item \textbf{object\_direction\_facing\_simple.}~Given three objects, after determining the perspective through two of them, determine the relative position of the other object in the video. Relative position includes left and right.
    \item \textbf{object\_volume\_comparison.}~Compare the volume sizes of multiple objects in the video to find the largest or smallest one.
    \item \textbf{object\_in\_frame}~Identify all the objects that appear in the video at a certain moment and output them by category.
    \item \textbf{object\_volume\_estimation.}~Identify the volume of the object and output the corresponding value.
    \item \textbf{object\_height\_determination.}~Identify the height differences of objects and determine which one is the highest.
    \item \textbf{temporal\_appearance\_sequence.}~Analyze the sequence of objects based on the chronological order of their appearance.
    \item \textbf{object\_nearby.}~For problems like identifying which objects are within 5 meters in front of a certain object. 
    \item \textbf{object\_below.}~Determine whether one object is positioned at a lower elevation relative to the other.
    \item \textbf{object\_above.}~Assess the relative vertical positions of two objects to establish whether one is positioned above the other.
    \item \textbf{high\_and\_low\_position.}~Given two objects, assess their relative vertical positions to identify which one is at a higher elevation.
\end{itemize}

\begin{figure*}[t!]
    \centering 
    \includegraphics[width=1\linewidth]{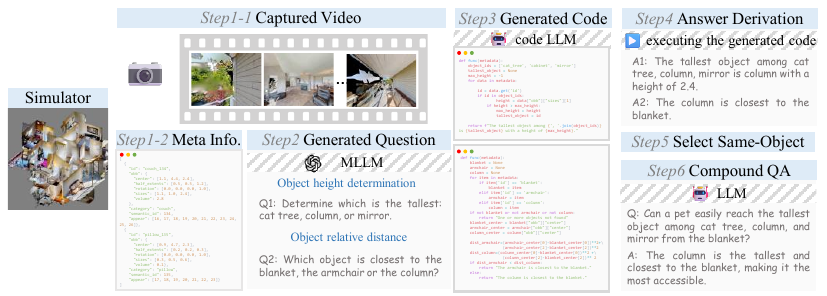} 
    \caption{Case Study} 
    \label{fig:case_study}
\end{figure*}

\subsection{Case Study}

We use a video as an example to demonstrate the process of data construction. As shown in Figure~\ref{fig:case_study}, we first obtain the meta-information in the video and the scene from the simulator. Then, based on the obtained meta-information and the images from the continuous frame extraction of the video, we generate the corresponding questions. For each problem, generate a piece of code for meta-information processing through a code LLM. The result of code execution is the ground truth. For the problem that meets the requirements, select two objects involving the same category and, through the reasoning ability of the large model, infer a new problem and its answer.

\subsection{Training Details}\label{hyperparameter}
To validate the effectiveness of our data, we trained on qwenvl-2.5-7B and InternVL-8B. The training parameters are shown in Table \ref{tab:training_configs}. We conducted the training on 8*A100 80G GPUs.

\begin{table}[t!]
  \centering
    \small 
  \caption{Training configurations for different models.} 
  \label{tab:training_configs} 
   \setlength{\tabcolsep}{5pt} 
  \begin{tabular}{lcc}
    \toprule
    \textbf{Configuration} & \textbf{Qwen2.5-VL-7B} & \textbf{InternVL-8B} \\
    \midrule
    SFT Method       & LoRA  & LoRA   \\
    Rank             & 256  & 256    \\
    Alpha            & 512   & 512   \\
    Epochs           & 1   & 1     \\
    Batch Size       & 2   & 2      \\
    Grad. Accum. Steps & 4   & 4  \\
    Learning Rate    & $5\times10^{-6}$  & $5\times10^{-6}$\\
    LR Scheduler     & Cosine  & Cosine\\
    \bottomrule
  \end{tabular}

\end{table}

\subsection{GRPO Training Details}\label{grpo_setting}
In our study, we employed reinforcement learning using the GPRO framework. A dataset of 20,000 samples was constructed and subsequently transformed into multiple-choice questions via GPT-4o. The reward function was designed to incorporate both rule-based rewards, derived from answer correctness, and format-related rewards. The model was trained on 8 A100 80GB GPUs with a batch size of 32, and each sample was rolled out 8 times during training. After 200 training steps, the reward function had converged, leading to the selection of 6.4k samples for comparative analysis.

\subsection{Benchmarks Description}\label{benchmark}

\begin{itemize}
    \item \textbf{VSI Bench.}~VSI-bench is an evaluation benchmark focused on video spatial understanding intelligence. It contains 5k question-and-answer pairs derived from 288 videos. It assesses a model's ability to understand video spatial information from several perspectives, including object spatial layout, object spatial attributes, and the temporal information of objects within a video.
    \item \textbf{ViewSpatial-Bench.}~ViewSpatial-bench is a multi-viewpoint spatial understanding benchmark featuring over 5.7 thousand question-answer pairs. Its tasks are divided into two main categories—Camera-based tasks and Person-based tasks—which collectively evaluate a model's spatial reasoning capabilities across various viewpoints and contexts. The former focuses on relative direction and object view orientation from a camera's perspective, while the latter includes object view orientation, relative direction, and scene simulation relative direction relative to people.
    \item \textbf{Q-Spatial-Scannet\&Q-Spatial-Plus.}~Q-Spatial-Bench is a benchmark designed to measure quantitative spatial reasoning. It involves quantitative metrics such as the distance and size between objects in the scene. Q-Spatial-Scannet is derived from the ScanNet annotation data, which is obtained via RGB-D scanning in a high-quality real-world environment. Q-Spatial-Plus is obtained from the actual measurement of real-world objects.
    \item \textbf{ERQA.}~ERQA is a benchmark for evaluating the embodied reasoning ability of multimodal models, including spatial reasoning and trajectory reasoning, action reasoning, state estimation, pointing, multi-view reasoning, and task reasoning.
    \item \textbf{CVBench.}~CVBench is a vision-centric test benchmark that includes 2,638 manually annotated examples. It covers a range of tasks, including 2D spatial relationship judgment, 2D counting, 3D depth prediction, and 3D relative distance prediction.
\end{itemize}

\subsection{Prompts} \label{spatial_prompt}
There are prompts to construct related questions and obtain ground truth. 

\begin{tcolorbox}[colback=gray!5!white,colframe=gray!75!black,title=generate image question prompt,breakable] \label{image_question_prompt}
\footnotesize\ttfamily
You are a teacher for an embodied task course, and your task is to create some test questions for an image-based based on the corresponding task type. \\
\#\#\#Noted \\
1.The proposed test questions must only use the candidate object information provided by the user, especially the object names, which must correspond one-to-one. The candidate objects include the object names and their corresponding category information\\
This is the name of the candidate object and the corresponding type of the candidate object, The content is a dictionary, with the key being the object name and the value being the object type.\\
\#\#\#candidate object \\
``` \\
\{objects\_info\}\\
```\\
2.The proposed test questions require that they can be solved through code based on the meta-information of the object. \\
The following is an example of meta-information. The object information in the example is not a candidate object.\\
\#\#\#meta information  \\
```\\
\{meta\_example\} \\
```\\
The content and meaning are as follows: \\
id: The unique name of an object \\
obb:obb is a three-dimensional oriented bounding box. In the coordinate system, the Y-axis is perpendicular to the ground and upward \\
    In obb information, "center" represents the centroid coordinate of an object in the world coordinate system \\
    "half\_extent" is the half-length of the OBB, indicating the distance from the center point to each face \\
    "sizes" is the size of the OBB, indicating the length, width, and height of the OBB. \\
    "volume" represents the size of the space occupied by OBB \\
"category" represents the type information of an object \\
In coordinate information, a movement of 0.1 unit is equivalent to 0.1 meter in reality \\
Additionally, the known extra information is the camera position: \\
camera\_position = [-1.703, 0.985824, 0.922993] \\
The raised questions need to comply with the following task types and indicate the task types in the output results. \\
```\\
\{question\_type\_all\} \\
```\\
4.The proposed test questions need to explicitly include the name information of the object and the name of the task type.\\
5.The output needs to be in JSON format. The output example is as follows, and the question types can refer to the example: \\
```\\
\{output\_example\} \\
```\\
instruction: Test question content \\
objects: The names of the objects involved in the question are selected from the names of the candidate objects \\
objcets\_category: what should be written here is the name of the object category
category: Task type \\
6.For numerical problems, units need to be included in the problem \\
7.For questions about object dimensions and the like, just ask about one dimension of length, width, or height in the question. \\
8.For a picture, please select the appropriate question type and ask at least 20 questions, and the number of questions of different types should be as even as possible. \\
9.Please refer to the given picture and propose questions with practical significance. \\
\end{tcolorbox}

\begin{tcolorbox}[colback=gray!5!white,colframe=gray!75!black,title=generate video question prompt,breakable]
\footnotesize\ttfamily
There is currently a first-person video centered on the camera. \\
Please create some test questions for the tasks in the embodied scenarios based on this video. \\
\#\#\#Noted \\
1.The proposed test questions can only use the candidate object information provided by the user, especially the object name information needs to correspond one-to-one\\
This is the name of the candidate object and the corresponding type of the candidate object \\ 
\#\#\# candidate object \\
``` \\
\{objects\_info\} \\
``` \\
2.The proposed test questions require that they can be solved through code based on the meta-information of the object.\\
\#\#\#meta information \\
```\\
\{meta\_example\} \\
```\\
The content and meaning are as follows \\
id: The unique name of an object \\
obb:obb is a three-dimensional oriented bounding box. In the coordinate system, the Y-axis is perpendicular to the ground and upward \\ 
    In obb information, "center" represents the centroid coordinate of an object in the world coordinate system \\
    "half\_extent" is the half-length of the OBB, indicating the distance from the center point to each face \\
    "rotation" is a quaternion representing the rotation of the OBB. \\
    "sizes" is the size of the OBB, indicating the length, width, and height of the OBB. \\
    "volume" represents the size of the space occupied by OBB \\
"category" represents the type information of an object \\
"appear" indicates in which pictures an object appears. For example, "appear = [0,1]" means it appears in the first and second pictures. The sequence of the pictures indicates the time information. \\
In coordinate information, a movement of 0.1 unit is equivalent to 0.1 meter in reality \\
3.The raised questions need to comply with the following task types and indicate the task types in the output results. \\
``` \\
\{question\_type\_all\} \\
``` \\ 
4.The proposed test questions need to explicitly include the name information of the object and the name of the task type.
5.The output needs to be in JSON format. The output example is as follows, and the question types can refer to the example:\\
```\\
\{output\_example\}\\
```\\
instruction: Test question content\\
objcets: The names of the objects involved in the test questions, and the names of the objects involved must explicitly appear in the instruction. \\
objcets\_category: what should be written here is the name of the object category
category: Task type\\
6.For a video, please select an appropriate type of question. At least 40 questions should be raised, and the number of questions of different types should be as even as possible.\\
7.For questions about object dimensions and the like, just ask about one dimension of length, width, or height in the question.\\
8.The perspective position information in the video is unknown, so the perspective information should not be assumed in the test questions. When designing test questions about perspective transformation, three objects need to be included to determine their relative positions.\\
9.Please refer to the given video and propose a video with practical significance.\\
\end{tcolorbox}

\begin{tcolorbox}[colback=gray!5!white,colframe=gray!75!black,title=generate image code prompt,breakable]
\footnotesize\ttfamily
You are a senior engineer of Python code. Your task is to write a function based on user questions, and the requirement is that the return result of the function is a string.\\
\#\#\#Known information\\
1.The function name is func, and the parameters pass a metadata variable and a camera\_position variable. The function is defined as follows \\
``` \\
def func(metadata,camera\_position):\\
``` \\
2. The content of metadata is a list. The following is part of the list. The content format is as follows. \\
Please refer to the format. The specific content is passed by the metadata variable.\\
```\\ 
\{meta\_info\} \\
```\\
The content and meaning are as follows:\\
id: The unique name of an object\\
obb: obb is a three-dimensional oriented bounding box. In the coordinate system, the Y-axis is perpendicular to the ground and upward. The system follows the right-handed coordinate rule.\\
    In obb information, "center" represents the centroid coordinate of an object in the world coordinate system\\
    "half\_extent" is the half-length of the OBB, indicating the distance from the center point to each face\\
    "sizes" is the size of the OBB, indicating the length, width, and height of the OBB.\\
    "volume" represents the size of the space occupied by OBB\\
"category" represents the type information of an object\\
In coordinate information, a movement of 0.1 unit is equivalent to 0.1 meter in reality\\
3.The content of camera\_position is a list. The content format is as follows. 
camera\_position =  [-1.703,0.985824,0.922993]\\
\#\#\#Noted \\
1.In the coordinate system, the Y-axis is perpendicular to the ground and upward. The system follows the right-handed coordinate rule.\\
2.The code part in the answer only needs to include the corresponding functions, and there is no need to provide calling examples.\\
3.Please use the object names involved in the question when completing the code.\\
4.If it is an object-counting problem, what is counted is the number of objects of the same type, and the category of objects needs to be used\\
5.The return result of the function must be a string and conform to the description in natural language\\
6.The output format can be referred to the reference code.\\
\#\#\#reference code. \\
```\\
\{reference\_code\} \\
```\\
\#\#\#question \\
```\\
\{question\} \\
```\\
\#\#\#The names of the objects involved in the question \\
```\\
\{objects\}\\
```\\
\#\#\#The category of objects involved in the problem\\
```\\
\{categories\} \\
```\\
\end{tcolorbox}

\begin{tcolorbox}[colback=gray!5!white,colframe=gray!75!black,title=generate video code prompt,breakable ]
\footnotesize\ttfamily
You are a senior engineer of Python code. Your task is to write a function based on user questions, and the requirement is that the return result of the function is a string.\\
\#\#\#Known information\\
1.The function name is func, and the parameter passes a metadata variable. The function is defined as follows\\
```\\
def func(metadata):\\
```\\
2.The content of metadata is a list. The following is part of the list. The content format is as follows. \\
Please refer to the format. The specific content is passed by the metadata variable. \\
```\\
\{meta\_info\}\\ 
```\\
The content and meaning are as follows\\
id: The unique name of an object\\
obb:obb is a three-dimensional oriented bounding box. In the coordinate system, the Y-axis is perpendicular to the ground and upward\\
    In obb information, "center" represents the centroid coordinate of an object in the world coordinate system\\
    "half\_extent" is the half-length of the OBB, indicating the distance from the center point to each face\\
    "rotation" is a quaternion representing the rotation of the OBB.\\
    "sizes" is the size of the OBB, indicating the length, width, and height of the OBB.\\
    "volume" represents the size of the space occupied by OBB\\
"category" represents the type information of an object\\
"appear" indicates in which pictures an object appears. For example, "appear = [0,1]" means it appears in the first and second pictures. The sequence of the pictures indicates the time information.\\
In coordinate information, a movement of 0.1 unit is equivalent to 0.1 meter in reality\\
\#\#\#Noted\\
1.The code part in the answer only needs to include the corresponding functions, and there is no need to provide calling examples.\\
2.Please use the object names involved in the question when completing the code.
3.If it is a object\_counting problem, what is counted is the number of objects of the same type, and the category of objects needs to be used\\
4.The output format and code logic can be referred to the reference code.\\
\#\#\#reference code. \\
```\\
\{reference\_code\} \\
```\\
\#\#\#question\\
```\\
\{question\}\\
```\\
\#\#\#The names of the objects involved in the question\\
```\\
\{objects\}\\
```\\
\#\#\#The category of objects involved in the problem\\
```\\
\{categories\}\\
```\\
\end{tcolorbox}

\end{document}